\documentclass[10pt,twocolumn,letterpaper]{article}

\usepackage{iccv}              % To produce the CAMERA-READY version
\usepackage{multirow}
\usepackage{multicol}
\usepackage{graphicx}
\usepackage{xcolor, colortbl}
\usepackage{bbm}
\definecolor{iccvblue}{rgb}{0.21,0.49,0.74}
\usepackage[pagebackref,breaklinks,colorlinks,allcolors=iccvblue]{hyperref}

\def\paperID{13055} % *** Enter the Paper ID here
\def\confName{ICCV}
\def\confYear{2025}

\definecolor{zk_fix}{HTML}{F26A4D}

\title{AVAM: Universal Training-free Adaptive Visual Anchoring Embedded into Multimodal Large Language Model for Multi-image Question Answering}

\author{
{Kang Zeng$^{1}$\footnotemark[1] \quad Guojin Zhong$^{1}$\footnotemark[1] \quad Jintao Cheng$^{2}$ \quad Jin Yuan$^{1}$\footnotemark[2] \quad Zhiyong Li$^{1}$\footnotemark[2]}\\
$^{1}$Hunan University\quad
$^{2}$South China Normal University\\
{\tt\small\{zengkang,gjzhong,yuanjin,zhiyong.li\}@hnu.edu.cn, 20172332035@m.scnu.edu.cn}\\
}
\begin{document}
\maketitle

\renewcommand{\thefootnote}{\fnsymbol{footnote}} 
\footnotetext[1]{Equal contribution.}
\footnotetext[2]{Corresponding authors.}

\begin{abstract}
The advancement of Multimodal Large Language Models (MLLMs) has driven significant progress in Visual Question Answering (VQA), evolving from Single to Multi Image VQA (MVQA). However, the increased number of images in MVQA inevitably introduces substantial visual redundancy that is irrelevant to question answering, negatively impacting both accuracy and efficiency. To address this issue, existing methods lack flexibility in controlling the number of compressed visual tokens and tend to produce discrete visual fragments, which hinder MLLMs' ability to comprehend images holistically. In this paper, we propose a straightforward yet universal Adaptive Visual Anchoring strategy, which can be seamlessly integrated into existing MLLMs, offering significant accuracy improvements through adaptive compression. Meanwhile, to balance the results derived from both global and compressed visual input, we further introduce a novel collaborative decoding mechanism, enabling optimal performance. Extensive experiments validate the effectiveness of our method, demonstrating consistent performance improvements across various MLLMs. The code will be publicly available.
\end{abstract}

\section{introduction}
Recently, Multimodal Large Language Models (MLLMs) have demonstrated remarkable cross-modal understanding and reasoning capabilities across various vision-language tasks \cite{alayrac2022flamingo, chen2024internvl2, li2023blip, liu2024improved, achiam2023gpt4}, significantly advancing Single-Image Visual Question Answering (SVQA) \cite{wang2024cogvlm, chen2023minigpt, chen2024sharegpt4v}. In this task, an MLLM processes a single image alongside natural language questions to generate accurate and contextually relevant answers. While the integration of visual information enhances the model’s capacity to predict answers by providing richer multimodal context, it also introduces computational inefficiencies in QA process. Additionally, redundant visual information may interfere with reasoning, potentially reducing the accuracy of the generated answers.

To address this challenge, existing SVQA approaches primarily adopt attention-based strategies \cite{yang2024visionzip, chen2025fastv, zhang2024sparsevlm} to highlight salient visual tokens or query-based strategies \cite{bai2023qwen, alayrac2022flamingo, laurenccon2024matters} to retrieve relevant visual tokens using a learnable query. These methods effectively reduce redundant visual tokens processed by MLLMs, enhancing the efficiency of the QA process while achieving answer accuracy comparable to or even higher than that of the original MLLMs.

\begin{figure}
    \centering
    \includegraphics[width=0.88\linewidth]{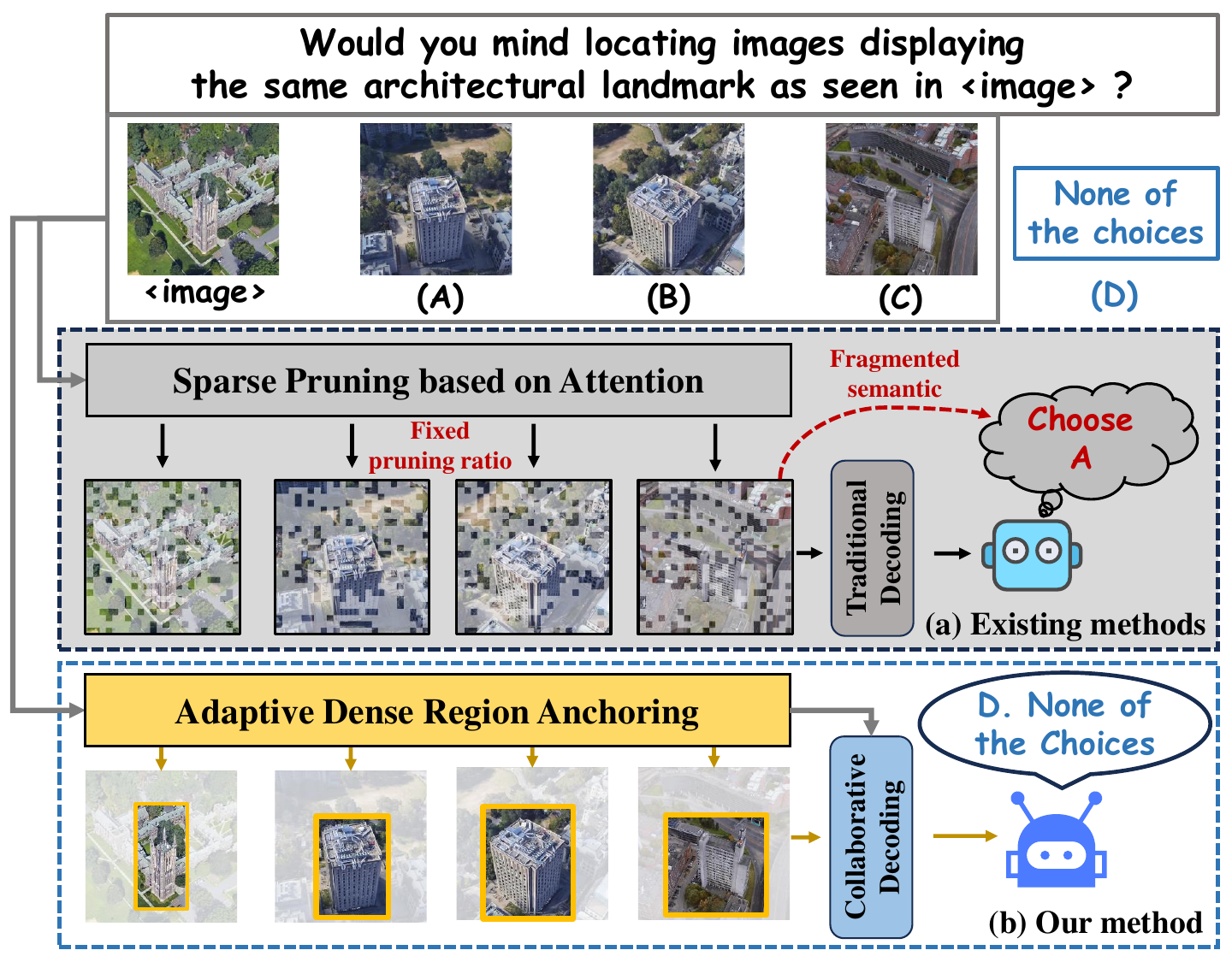}
    \caption{An example to illustrate the differences in visual pruning between our approach and other methods. While existing methods rely on a fixed number of visual tokens, leading to fragmented visual representations, our method adaptively prunes continuous key regions across multiple images.}
    \label{fig: intro}
\vskip-3ex
\end{figure}

Research attention toward Multi-Image Visual Question Answering (MVQA) has increased significantly due to its practical relevance.
Unlike SVQA, MVQA involves injecting multiple images into MLLMs for question answering, thereby furnishing abundant semantics through temporal/spatial views to facilitate complex reasoning~\cite{wang2024muirbench,liu2024mibench}. This mode, where multimodal data with interleaved image-text interactions demonstrates more pronounced practical utility, enables applications like event prediction via temporal image sequences~\cite{fu2025videomme,li2024mvbench,zhou2025mlvu} and enhanced perceptual analysis through spatial multi-image integration~\cite{qian2024nuscenesQA,sima2024drivelm,wu2025nuprompt}.
% However, visual redundancy becomes more pronounced in multi-image paradigm, where key visual information essential for reasoning is increasingly obscured as the number of images expands. Existing visual redundancy filtering approaches in SVQA encounter three major challenges when applied to MVQA. 
% First, they rely heavily on task-specific fine-tuning with training data and struggle to strike a balance between accuracy and efficiency. 
% Second, most methods tend to enforce an excessively high compression ratio, leading to sparse selection of key tokens. This disrupts the spatial continuity and local semantics within the image (as illustrated in Fig.~\ref{fig: intro}(a)). This problem is further amplified in multi-image scenarios, where sparse token selections from individual images can interfere with cross-image correlation modeling and undermine spatial consistency across frames. 
% Third, these methods require manually setting a fixed redundancy filtering ratio as a hyperparameter. However, since the proportion of visual redundancy varies across different images, this fixed setting may lead to the loss of critical information or excessive retention of redundant data.
However, visual redundancy becomes more pronounced in multi-image paradigm, where key visual information is increasingly obscured as the number of images expands. Existing visual redundancy filtering approaches encounter three major challenges when applied to MVQA:
First, learning-based approaches~\cite{alayrac2022flamingo,li2023blip,bai2023qwen,cha2024honeybee,dai2023instructblip}, represented by those utilizing learnable queries to extract specific deep-level visual features akin to using a mold for forging, often require substantial training data to develop such a generalizable ``feature mold''. This data dependency and inherent scalability constraints hinder optimal accuracy-efficiency trade-offs.
Second, many training-free methods~\cite{chen2025fastv,zhang2024sparsevlm,zhang2024fastvlm,yang2024visionzip} typically mandate fixed filtering ratios as hyperparameters. Given varying redundancy levels across images, this rigidity causes critical information loss or redundant data retention.
Third, another challenge prevalent in training-free methods stems from their discrete visual token selection process. This approach disrupts the spatial continuity and local semantics within an image (as illustrated in Fig.~\ref{fig: intro}(a)). Crucially, this problem is amplified in multi-image scenarios, sparse and discrete token selections from individual images can interfere with modeling cross-image correlations and undermine spatial consistency across frames.

To address the aforementioned challenges and enhance MLLM's capability in processing multi-image inputs, this paper proposes a training-free, adaptive visual anchoring strategy that extracts critical regions from global visual representations through the response map quantifying cross-modal relevance between visual features and corresponding text. Based on response hotspots within this map, we generate a series of anchor boxes to localize question-relevant visual regions, subsequently introducing a response density metric to preserve the most critical region and filter other redundant visual tokens.
Furthermore, we propose a novel collaborative decoding mechanism that dynamically fuses decoding distributions under global visual context according to varying redundancy levels, harmonizes responses from both global and compressed local visual cues and achieves improved performance.

The main contributions of this work can be summarized:
\begin{itemize}
    \item We propose an adaptive visual anchoring strategy tailored for MVQA, which serves as a training-free pipeline that can be seamlessly integrated into polular MLLMs. To the best of our knowledge, this is the first work to systematically explore the issue of visual redundancy in MVQA.
    \item We propose a novel collaborative decoding mechanism that dynamically balances probability distributions between global visual context and local cues according to varying visual redundancy ratios.
    \item Extensive experiments demonstrate that our method can be effectively extended to most popular MLLMs, enhancing their performance on MVQA.
\end{itemize}

\section{Related Work}
\subsection{Multimodel Large Language Models}
Driven by the remarkable breakthroughs of LLMs in natural language understanding and generation \cite{dubey2024llama, zheng2023vicuna, glm2024chatglm, achiam2023gpt4, liu2024deepseekv3}, MLLMs extended from LLMs with visual capabilities have demonstrated impressive performance in multimodal understanding \cite{wang2024qwen2,team2023gemini,achiam2023gpt4,ye2024mplug3,liu2024visual}. Existing MLLMs \cite{liu2024visual,lu2024deepseek,jiang2024mantis} typically convert images into visual token sequences via the patch embedding process of pre-trained vision encoders. For instance, LLaVA-v1.5 \cite{liu2024improved} transforms a $336 \times 336$ resolution image into $576$ visual tokens. In multi-image scenarios, the number of visual tokens could exceed text tokens by hundreds of times. However, most current MLLMs store visual semantics through fixed quantities of visual tokens per image, failing to dynamically extract critical visual features for downstream tasks. This leads to substantial retention of task-irrelevant redundant visual information when fed into LLM, consequently impairing the performance.

\subsection{Visual Token Compression for VQA}
Compared with text tokens, visual tokens exhibit strong sparsity in semantic representation, revealing significant compression potential. MLLMs based on learnable visual token compression strategies~\cite{bai2023qwen,laurenccon2024matters,alayrac2022flamingo} employ learnable queries to abstract visual features, substantially reducing visual token counts and achieve a strong performance on VQA tasks. For example, Qwen-VL \cite{bai2023qwen} compresses visual tokens per image from $1024$ to $256$ via a cross-attention adapter. Meanwhile, training-free methods have attracted considerable research interest due to their minimal computational cost and flexibility. Recent works \cite{chen2025fastv} first investigated inefficient visual attention patterns in popular MLLMs and pruned visual tokens based on attention scores. Follow-up studies \cite{zhang2024fastvlm, zhang2024TokenCorrCompressor,yang2024visionzip} eliminate redundancy by selecting dominant visual tokens through internal correlations. Advanced approaches \cite{zhang2024sparsevlm, han2024freeVideoLLM} incorporate textual information for cross-modal token pruning. Notably, \cite{liu2024MustDrop} proposed a multi-stage dynamic optimization framework considering varying importance levels of visual tokens across different lifecycle stages. Distinct from these methods, this work focuses not on pursuing extreme compression ratios, but rather on dynamically compressing visual tokens according to conversational contexts to maximally preserve visual information beneficial for the final responses.

\section{Preliminaries}
\subsection{Task definition}
\label{preliminaries}
Multi-image Visual Question Answering (MVQA) involves providing a set of $N$ image prompts $\mathcal{X}=\{x_0, x_1,...,x_{N-1}\}$ and a text prompt $\mathcal{T}$ as inputs to a Vision-Language Model (VLM) to generate a textual answer $\mathcal{A}$. Typically, the text prompt $\mathcal{T}$ corresponds to a question text $\mathcal{Q}_t$, i.e., $\mathcal{T}=\mathcal{Q}_t$, and may include additional textual supplements $\mathcal{C}=\{c_0, c_1,...,c_{N-1}\} \subseteq \mathcal{T}$, which are associated with each image in $\mathcal{X}$. These supplements can provide richer knowledge about the images, aiding VLMs in generating accurate and contextually relevant answers.

To bridge the representation gap between visual and textual input for MVQA, it is essential to generate consistent embeddings for both modalities. Specifically, the VLM encodes the textual prompt $\mathcal{T}$ into text embeddings $\mathcal{H}_T=\mathcal{E}_T(\mathcal{T}) \in \mathbb{R}^{L \times D}$ using a text encoder $\mathcal{E}_T$, where $L$ represents the number of textual tokens and $D$ denotes the embedding dimension of each token. Simultaneously, a vision encoder $\mathcal{E}_I$ extracts visual features from the image prompts, which are then projected into the language space $\mathcal{H}_V \in \mathbb{R}^{N \times K \times D}$ via a vision-to-text projector $\mathcal{P}$:
\begin{equation}
    \mathcal{H}_V = \mathcal{P}(\mathcal{E}_I(\mathcal{X})),
\end{equation}
where $K=\frac{H}{P} \times \frac{W}{P}$ denotes the number of visual tokens per image, typically determined during the patch embedding stage of the vision encoder. Here, $H$ and $W$ represent the image height and width, respectively, while $P$ is the patch size. Finally, the visual features $\mathcal{H}_V$ are aligned with their corresponding textual features $\mathcal{H}_T$. 
% In certain tasks, an additional query image $\mathcal{Q}_i$ is introduced as part of the question. This query image is encoded alongside the other image prompts, resulting in visual embeddings $\mathcal{H}_V \in \mathbb{R}^{(N+1) \times K \times D}$. 
These aligned embeddings are then combined and injected into a large language model (LLM) to generate the answer $\mathcal{A}$:
\begin{equation}
    \mathcal{A}=LLM([H_V, H_T]).
    \label{llm}
\end{equation}

%\mathcal{X}=\{x_0, x_1, ...,x_{N-1},\mathcal{Q}_i\}$. 

\subsection{Submergence of critical visual tokens}
\label{submergence}
\begin{figure}
    \centering
    \includegraphics[width=0.9\linewidth]
    {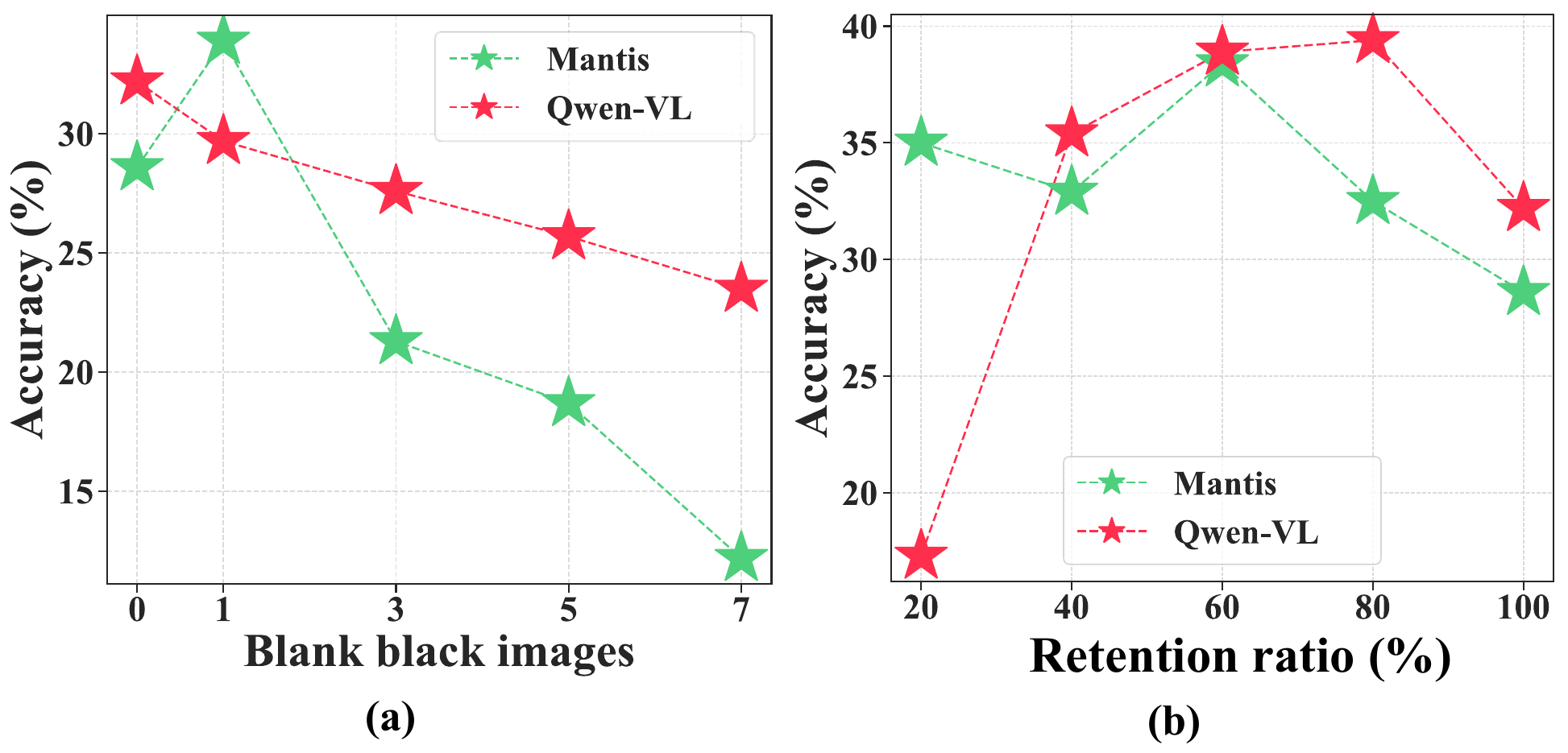}
    \caption{Submergence of critical visual tokens. Subfigure (a) directly demonstrates the submergence of critical visual tokens as redundant images increase, while subfigure (b) indirectly validates this phenomenon through the refinement of critical visual tokens.}
    \label{fig:analysis}
\vskip-3ex
\end{figure}

Our central insight is that as the number of input visual tokens increases, certain tokens highly related to the question may become obscured, yielding incorrect answers by the MLLM. To verify this hypothesis, we conduct two testings. First, we continuously add blank black images to the original image prompts $\mathcal{X}$
while keeping the question unchanged. We select Mantis \cite{jiang2024mantis} as representatives of insertion-based methods, and Qwen-VL \cite{bai2023qwen} as representative of query-based methods. Utilizing the Fine-grained Visual Recognition task in MIBench~\cite{liu2024mibench}, we perform a quantitative analysis to explore the relationship between image redundancy and MLLM performance. As depicted in Fig.~\ref{fig:analysis} (a), the accuracy of Qwen-VL decreases with increasing number of redundant images, while Mantis exhibits performance fluctuations after the introduction of a single redundant image, subsequently showing a similar pattern of accuracy degradation. Meanwhile, the accuracy of Mantis shows a significant downward trend, whereas the decline for Qwen-VL is more gradual. We attribute this primarily to Qwen-VL's use of a learning-based visual token compression method, which partially mitigates the submergence of critical visual tokens. Notably, the performance decline may also be associated with the distribution shift induced by introducing black images, and we provide the analysis in the supplementary materials~\ref{sup_the_impact_of_black_images}.

Second, we attempt to filter out redundant visual tokens to assist MLLMs in identifying question-critical ones. Following the approach proposed in \cite{han2024freeVideoLLM}, we calculate the cosine similarity between each image feature and its corresponding caption, retaining the top $R$\% of visual tokens most relevant to the text as input features for the LLM. As shown in Fig. \ref{fig:analysis} (b), when the token retention ratio $R$ decreases from $100$\% to $20\%$, the answer accuracy of the MLLM response initially improves and then declines. This observation suggests that critical visual tokens can become overshadowed by irrelevant information, and refining these tokens could enhance the MLLM's performance.

\section{Method}
\begin{figure*}
    \centering
    \includegraphics[width=0.88\linewidth]{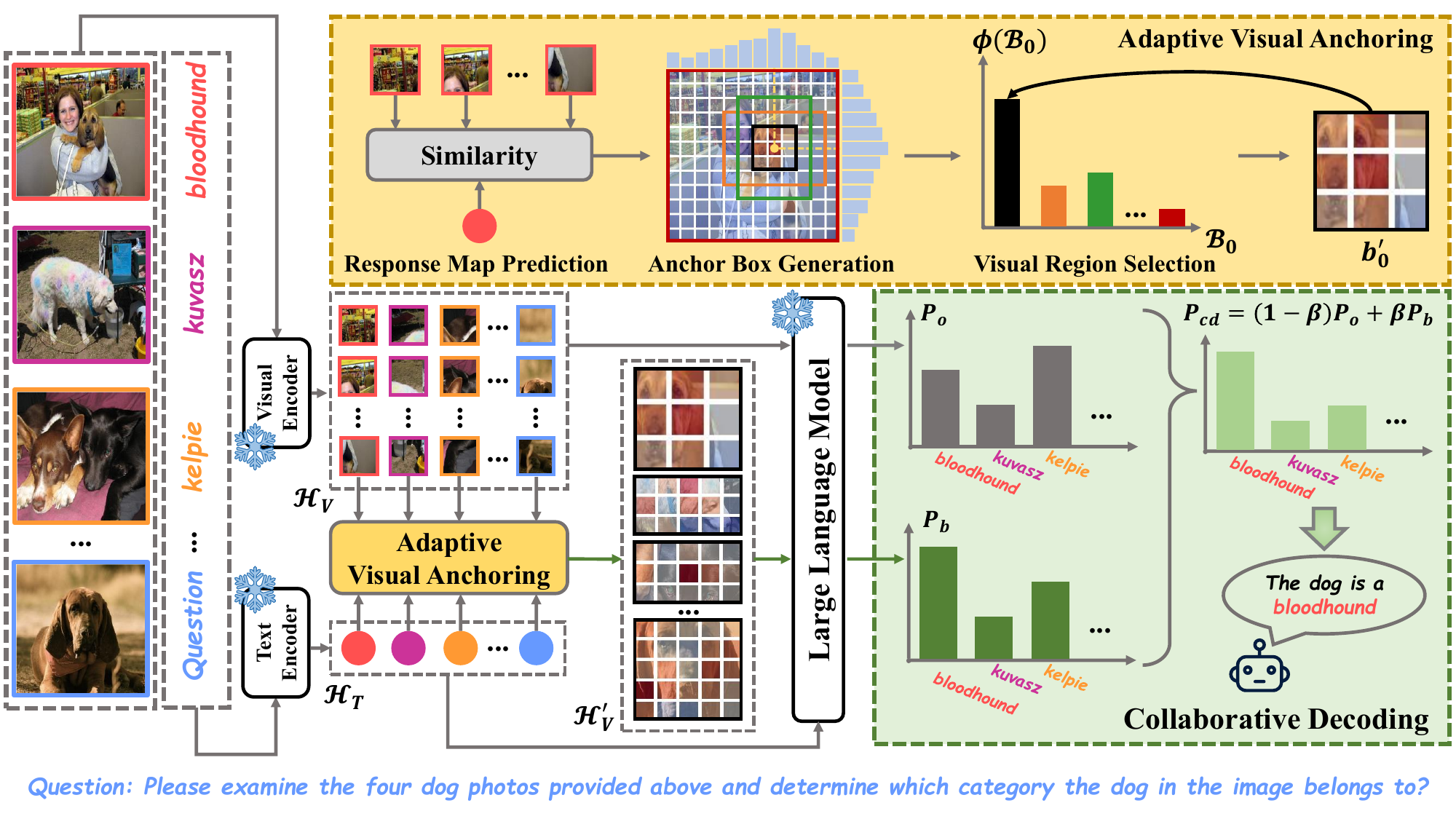}
    % \caption{Workflow of our proposed method. Critical visual regions are obtained through implicit retrieval between the visual features of each image and their corresponding caption embeddings. In the example shown in the figure, the visual region corresponding to the dog, as aligned with the text, is successfully retrieved. Gaussian noise is added to the critical visual regions within the global visual features to serve as the distorted input for GICD. Finally, the probability distribution of the next token is obtained through contrastive decoding, leveraging the probability distributions derived from the original input $P_o$ and the distorted input $P_n$.}
    \caption{Workflow of our proposed method. Critical visual regions are obtained through implicit retrieval between the visual features of each image and their corresponding caption embeddings. In the example shown in the figure, the visual region corresponding to the dog, as aligned with the text, is successfully retrieved. Finally, the probability distribution of the next token is obtained through collaborative decoding, leveraging the probability distributions derived from the original input $P_o$ and the critical input $P_b$.}
    \label{fig:architecture}
\end{figure*}

% \subsection{Adaptive visual anchoring with text-image response}
\subsection{Adaptive visual anchoring}
\label{sec41}
As analyzed in Section \ref{submergence}, the increasing number of images in MVQA introduces a significant amount of redundant visual tokens, which causes the essential tokens to be submerged. We propose an adaptive visual anchoring strategy that extracts the most critical region from the global image based on text-image responsiveness to address this issue. This approach effectively filters out query-irrelevant visual regions prior to feeding the image to the LLM. (see Eq.~\eqref{llm}), providing concise and relevant prompts for question answering. The process of our proposed approach is illustrated in Fig. ~\ref{fig:architecture}, which consists of three sequential steps: token-level response map prediction, hotspot-centered anchor box generation, and optimal visual region selection.

%Given the visual representation $\mathcal{H}_V$ and textual representation $\mathcal{H}_T$, our method adaptively selects the most relevant visual region $b_i^\prime$ inside each image $x_i$ to replace the entire image as input to the LLM. Specifically, the proposed approach is composed of three sequential steps (see Figure \ref{}):
\subsubsection{Token-level response map prediction}
\label{sec411}
To distinguish between relevant and irrelevant visual tokens in an image, this step aims to compute a response score for each token, reflecting its relevance to the text prompt. Specifically, given visual tokens $V_k$ represented by $\mathcal{H}_{V_k} \in \mathbb{R}^D$ in an image $x$, and the corresponding text prompt embedding $\mathcal{H}_{T_x} \in \mathbb{R}^{L \times D}$ associated with $x$, our approach calculates the response value $\mathcal{S}_k$ of the visual token $V_k$ with respect to the text embedding using cosine similarity based on the alignment of visual and textual features.
\begin{equation}
\label{response_map}
    \mathcal{S}_k=\cos \left\langle\sigma(\mathcal{H}_{T _x}), \mathcal{H}_{V_k}\right\rangle,
\end{equation}
where $\sigma(\cdot)$ represents feature aggregation along the length dimension using global average pooling. Therefore, our approach generates a response map $\mathcal{S} \in \mathbb{R}^K$ for the image $x$, where each element $\mathcal{S}_k$ corresponds to a text-response score for the visual token $V_k$ relative to the global text description of $x$. 
% In cases where the image $x$ is not associated with a text prompt, our approach substitutes the question text $\mathcal{Q}_t$ for the text prompt. 
When image $x$ lacks a corresponding text prompt, our approach uses the question text $\mathcal{Q}_t$ as a substitute. In this case, $\mathcal{S}_k$ reflects the relevance of the visual token $V_k$ with respect to the question, enabling MLLMs to identify valuable visual regions that align with the question’s requirements. Regarding the effectiveness of the text selection strategy, we investigate it in our supplementary materials \ref{sup_text_selection}.

%To extract the most valuable visual tokens from each image, we define the notion of text-image response map $\mathcal{S}$, which measures the relevance of visual information to the query text. This responsiveness can be computed via cosine similarity, as shown in Equation \ref{response_map}. Specifically, the text features are first aggregated along the length dimension using global average pooling to obtain a global representation of the text. When the text prompt contains descriptions for each image as discussed in \ref{preliminaries}, the visual features of an image are compared against the embedding of its corresponding textual description. In contrast, when the text prompt only contains the question text, all images are calculated with the global text embedding of the question. This allows the model to identify valuable visual regions that align with the question's demand. 

\subsubsection{Hotspot-centered anchor box generation}
\label{sec412}
The token-level response map effectively captures the relevance of each token to the text prompt, providing a clear indication of the hotspot region in the image. However, as discussed in Section \ref{sec411}, the independent calculation of visual tokens inevitably leads to the issue of ``fragmented visual semantics''. This disruption of the inherent spatial relationships between visual tokens makes it difficult for the model to fully comprehend the semantics of the original image. To accurately capture the hotspot region, we design a hotspot-centered anchor box generation strategy. Notably, our approach first identifies the hotspot center based on the response map $\mathcal{S}$, considering the spatial correlation within the map. In detail, we reshape $\mathcal{S} \in \mathbb{R}^K$ into a 2D image format $\mathcal{S} \in \mathbb{R}^{U \times V}$, where $U=\frac{W}{P}$ and $V=\frac{H}{P}$ is the width and height of the 2d response map,
 and then calculate the hotspot center 
$(u_{c},v_{c})$ using the gravity center formula:
\begin{equation}
\label{centeroid}
 \begin{split}
 % u_{c}=\left\lfloor\frac{\sum\limits_{v} \sum\limits_{u}  \mathcal{S}(u, v) \times u}{\sum\limits_{v} \sum\limits_{u}  \mathcal{S}(u, v)}\right\rfloor, \\
 % v_{c}=\left\lfloor\frac{\sum\limits_{v} \sum\limits_{u}\mathcal{S}(u, v) \times v}{\sum\limits_{v} \sum\limits_{u}  \mathcal{S}(u, v)}\right\rfloor, 
 (u_c, v_c)=(\left\lfloor\frac{\sum\limits_{v} \sum\limits_{u}  \mathcal{S}(u, v) \times u}{\sum\limits_{v} \sum\limits_{u}  \mathcal{S}(u, v)}\right\rfloor, \left\lfloor\frac{\sum\limits_{v} \sum\limits_{u}\mathcal{S}(u, v) \times v}{\sum\limits_{v} \sum\limits_{u}  \mathcal{S}(u, v)}\right\rfloor)
 \end{split}
\end{equation}
% where $\mathcal{S}(u, v)$ represents the response value at position $(u,v)$ in $\mathcal{S}$. Next, our approach generates $M$ anchor boxes $\mathcal{B}=\{b_0,b_1,...,b_{M-1}\}$ centered at $(u_{c},v_{c})$, where the number of the anchor boxes are manually set in our experiments. Unlike YOLO, which requires anchor boxes for all image patches, our approach only places anchor boxes centered around one patch, thus minimizing the impact on retrieval speed.
where $\mathcal{S}(u, v)$ represents the response value at position $(u,v)$ in $\mathcal{S}$. Next, our method systematically generates all possible anchor boxes $\mathcal{B}=\{b_0,b_1,...,b_{M-1}\}$ centered at the point $(u_c, v_c)$ on the response map. Concretely, the initial anchor box is centered at $(u_c, v_c)$ with height $1$ and width $1$, and then it continuously expands in both height and width until it reaches the boundary of the image $\mathcal{S}$. As a result, multiple anchor boxes are generated by Equation~\eqref{anchor_box}.
% \begin{equation}
%     \label{anchor_box}
%     \begin{aligned}
%         w=\{1 + j \cdot & [\mathbbm{1}\left(u_{c}-j \geq 0 \right) \\
%         & + \mathbbm{1}\left(u_{c}+j < U\right)] \}_{j=0}^{max(u_{c}, U-u_{c})}, \\
%         h=\{1 + j \cdot & [\mathbbm{1}\left(v_{c}-j \geq 0\right) \\
%         & + \mathbbm{1}\left(v_{c}+j < V\right)] \}_{j=0}^{max(v_{c}, V-v_{c})}, \\
%     \end{aligned}
% \end{equation}
\begin{equation}
    \label{anchor_box}
    \left\{
    \begin{aligned}
        &w=\{1 + min(j, u_c) + min(j, U-u_c) \}_{j=0}^{max(u_{c}, U-u_{c})}, \\
        &h=\{1 + min(k, v_c) + min(k, V-v_c) \}_{k=0}^{max(v_{c}, V-v_{c})}, \\
        &\mathcal{B}=\{(u_c, v_c, w_j, h_k) \mid w_j \in w, h_k \in h\},
    \end{aligned}
    \right.
\end{equation} 

\subsubsection{Optimal visual region selection}
\label{sec413}
Among all the generated anchor boxes, we select the one that encapsulates the most relevant visual tokens as the optimal cropped region $b^\prime$, which serves as the input to the LLM for generating the final answer. To quantitatively evaluate the distribution of valuable visual tokens $p_j$ within each anchor box $b_m$, we define a response density function $\psi(b_m)$: 
\begin{equation}
\label{density}
    \psi(b_m)=\frac{\sum_{p_j \in b_m} \mathcal{S}\left(p_j\right)}{U_{b_m} \times V_{b_m}},
\end{equation}
where $V_{b_m}$ and $U_{b_m}$ denote the height and width of the anchor box, and $\mathcal{S}\left(p_j\right)$ is the response score of $p_j$. $\psi(b_m)$ reflects the average response value in the anchor box $b_m$, where the high value indicates the strong relevance of this anchor box with respect to the corresponding text. Therefore, our approach selects the anchor box with the highest response density as the final output $b^\prime$. 
% Notably, when an anchor box contains only a single visual token with the maximum response value, it naturally achieves the highest response density. To avoid such extreme cases, we introduce a minimum crop ratio $R$ as an additional constraint to guide this process, expressed as:
\begin{equation}
\label{optimal_select}
    b^\prime=\underset{b_m \in \mathcal{B} \wedge \frac{U_{b_m} \cdot V_{b_m}}{U \cdot V} \geq R}{\operatorname{argmax}} \psi(b_m),   
\end{equation}

\subsection{Collaborative decoding}
\label{42}
After identifying the key visual regions, we aim for MLLMs to simultaneously consider global visual semantics and local critical features when answering questions. In contrast to popular contrastive decoding methods \cite{m3id,wang2024valid}, which suppress adverse effects through adversarial interpolation between the decoding results of the original and distorted inputs, our proposed collaborative decoding dynamically weights the probability distributions derived from both the original visual prompt input and the key visual region input based on their redundancy levels.

Specifically, given the original text prompt $\mathcal{T}$, visual prompt $\mathcal{X}$, and the sequence of text tokens $\mathcal{A}_{<t}$ generated up to step $t-1$, we predict the probability distribution of the $t$-th text token $\mathcal{A}_{t}$ under the original input, denoted as $P_{o}\left(\mathcal{A}_t \mid \mathcal{T}, \mathcal{X}, \mathcal{A}_{<t}\right)$, through an autoregressive process. Similarly, when using the key visual regions $b^\prime$ extracted as the visual source, we obtain the corresponding probability distribution $P_{b}\left(\mathcal{A}_t \mid \mathcal{T}, b^\prime, \mathcal{A}_{<t}\right)$. The collaborative decoding process is represented in Equation \eqref{decoding}:
\begin{equation}
    \begin{aligned}
    \label{decoding}
        P_{cd}\left(\mathcal{A}_t \mid \mathcal{T}, \mathcal{X}, \mathcal{b}^\prime, \mathcal{A}_{<t}\right) & =(1-\beta) P_{o}\left(\mathcal{A}_t \mid \mathcal{T}, \mathcal{X}, \mathcal{A}_{<t}\right) \\
        & + \beta P_b\left(\mathcal{A}_t \mid \mathcal{T}, b^\prime, \mathcal{A}_{<t}\right),
    \end{aligned}
\end{equation}
where $\beta$ is the collaboration coefficient. A smaller value of $\beta$ indicates that MLLMs should focus more on the responses derived from the key visual regions. Subjectively, a smaller extracted key visual region indicates a higher level of redundancy in the image, increasing the likelihood that critical visual tokens may be overshadowed. Consequently, the collaborative effect should be diminished to suppress semantic distortion caused by redundancy, suggesting that the strength of collaboration is negatively correlated with visual redundancy. We set $\beta=e^{-\lambda r}$ to control the collaboration strength, where $\lambda$ is a hyperparameter and $r$ represents the visual redundancy rate calculated using Equation \eqref{redundancy}.
\begin{equation}
    \label{redundancy}
    r=1 - \frac{\sum_i^N U_i^\prime \times V_i^\prime}{N \times U \times V},
\end{equation}
where $U_i^\prime$ and $V_i^\prime$ represent width and height of the most critical region $b_i^\prime$ extracted from the $i$-th image, respectively.
\begin{table*}[!t]
\centering
\small
\resizebox{0.83\textwidth}{!}{
\begin{tabular}{l|ccccccccccccc}
\toprule
\multirow{2}{*}{\textbf{Model}} & \multicolumn{13}{c}{\textbf{MuirBech}} \\
 & \textbf{GU} &
\textbf{C} &
\textbf{AU} &
\textbf{VG} &
\textbf{ITM} &
\textbf{O} &
\textbf{SU} &
\textbf{DS} &
\textbf{CU} &
\textbf{DU} &
\textbf{AS} &
\textbf{VR} &
\textbf{Avg} \\
\midrule

LLaVA-v1.5-7B & \textbf{38.0} & 21.8 & 1.2 & 21.4 & 28.7 & 9.4 & 44.1 & \textbf{17.9} & 24.4 & \textbf{34.7} & 24.0 & \textbf{24.3} & 24.2 \\
LLaVA-v1.5-7B$^{\dag}$ & 34.0 & \textbf{23.1} & \textbf{31.1} & \textbf{22.6} & \textbf{31.0} & \textbf{14.1} & \textbf{50.0} & \textbf{17.9} & \textbf{25.6} & 30.7 & \textbf{29.1} & 21.6 & \textbf{27.6} \\

\hline
DeepSeek-VL-7B & 25.0 & 24.0 & 7.9 & 20.2 & 37.5 & \textbf{9.4} & \textbf{44.1} & 28.2 & \textbf{33.3} & \textbf{50.8} & \textbf{27.6} & \textbf{19.2} & 27.3 \\
DeepSeek-VL-7B$^{\dag}$ & \textbf{31.0} & \textbf{24.8} & \textbf{28.1} & \textbf{22.6} & \textbf{40.1} & 4.7 & 43.6 & \textbf{32.7} & \textbf{33.3} & \textbf{50.8} & 25.5 & 17.5 & \textbf{29.5} \\

\hline
Mantis-8B & \textbf{21.0} & 30.3 & 36.6 & \textbf{16.7} & 41.8 & \textbf{17.2} & 54.3 & \textbf{24.4} & \textbf{43.6} & 55.5 & 34.7 & 31.9 & 34.0 \\
Mantis-8B$^{\dag}$ & \textbf{21.0} & \textbf{30.8} & \textbf{37.2} & \textbf{16.7} & \textbf{44.0} & \textbf{17.2} & \textbf{55.4} & 23.2 & \textbf{43.6} & \textbf{55.8} &  \textbf{35.7} & \textbf{32.9} & \textbf{34.5}\\

\hline
InternVL2-8B & 30.0 & \textbf{29.1} & 32.3 & \textbf{34.5} & 46.6 & \textbf{9.4} & 52.2 & 27.4 & \textbf{39.7} & 46.5 & 49.5 & 41.8 & 36.6 \\
InternVL2-8B$^{\dag}$ & \textbf{42.0} & 27.4 & \textbf{37.2} & 33.3 & \textbf{51.3} & \textbf{9.4} & \textbf{57.0} & \textbf{33.2} & \textbf{39.7} & \textbf{49.0} & \textbf{50.0} & \textbf{43.8} & \textbf{39.5} \\

\hline
Qwen-VL-7B & \textbf{7.0} & 10.3 & \textbf{23.2} & \textbf{20.2} & 12.9 & 15.6 & \textbf{24.7} & 12.9 & 23.1 & 2.0 & 3.1 & 8.6 & 13.6 \\
Qwen-VL-7B$^{\dag}$ & \textbf{7.0} & \textbf{10.7} & 22.6 & 17.9 & \textbf{13.2} & \textbf{17.2} & 24.2 & \textbf{14.4} & \textbf{24.4} & \textbf{2.3} & \textbf{4.6} & \textbf{9.6} & \textbf{14.0}\\

\hline
Idefics2-8B & 21.0 & 21.8 & \textbf{28.7} & 23.8 & 26.9 & 15.6 & \textbf{55.4} & \textbf{24.4} & \textbf{39.7} & 19.1 & 18.9 & 18.2 & 26.1 \\
Idefics2-8B$^{\dag}$ & \textbf{33.0} & \textbf{29.1} & 25.0 & \textbf{25.0} & \textbf{29.0} & \textbf{21.9} & 39.8 & \textbf{24.4} & 35.9 & \textbf{36.7} & \textbf{23.5} & \textbf{29.1} & \textbf{29.4}\\

\hline
LLaVA-OV-Qwen2-7B & 42.0 & 27.8 & 37.8 & \textbf{28.6} & 46.1 & 14.1 & \textbf{67.7} & 25.3 & 34.6 & 56.5 & 38.8 & 45.9 & 38.8 \\
LLaVA-OV-Qwen2-7B$^{\dag}$ & \textbf{46.0} & \textbf{30.3} & \textbf{40.2} & 27.4 & \textbf{49.6} & \textbf{15.6} & 62.4 & \textbf{36.2} & \textbf{35.9} & \textbf{59.6} & \textbf{43.9} & \textbf{46.9} & \textbf{41.2} \\
% Qwen2-VL-7B & 80.6 & 78.5 & 31.7 & 58.5 & 70.3 & 63.9 & 95.4 & 78.4 & 39.6 & 59.5 & 68.2 \\

% Qwen2-VL-7B\dag &  &  &  &  &  &  &  &  &  &  &  \\ 

\bottomrule
\end{tabular}}
\caption{Performance comparison by integrating our compression method (noted as ``\dag'') into different MLLMs tested on the MuirBench. \textbf{Avg} indicates the average accuracy across the benchmark.}
\label{tab:res_on_muirbench}
% \vspace{-2mm}
\end{table*}
\begin{table*}[!t]
\centering
\small
\resizebox{0.83\textwidth}{!}{
\begin{tabular}{l|cccccc|ccccc|c}
\toprule
\multirow{2}{*}{\textbf{Model}} & \multicolumn{6}{c|}{\textbf{MIBench MII}} & \multicolumn{5}{c|}{\textbf{MIBench MKS}} & \textbf{Mantis} \\
 & \textbf{GC} &
\textbf{SD} &
\textbf{VR} &
\textbf{TR} &
\textbf{LR} &
\textbf{Avg} &
\textbf{FVR} &
\textbf{TRI} &
\textbf{VTK} &
\textbf{TVK} &
\textbf{Avg} & \textbf{Eval}\\
\midrule

LLaVA-v1.5-7B & 50.3 & 14.4 & 15.5 & 0.5 & 0.1 & 16.2 & 29.2 & \textbf{18.9} & \textbf{14.9} & 21.2 & 21.1 & 37.1 \\
LLaVA-v1.5-7B$^{\dag}$ & \textbf{54.6} & \textbf{15.6} & \textbf{16.5} & \textbf{30.1} & \textbf{41.6} & \textbf{31.7} & \textbf{36.8} & \textbf{18.9} & 11.1 & \textbf{22.2} & \textbf{22.3} & \textbf{39.3} \\
\hline
DeepSeek-VL-7B & \textbf{60.3} & 64.2 & 41.7 & 1.0 & 0.0 & 33.4 & 65.1 & 53.3 & \textbf{18.3} & 38.6 & 43.8 & 39.8 \\
DeepSeek-VL-7B$^{\dag}$ & 59.7 & \textbf{64.9} & \textbf{44.0} & \textbf{37.1} & \textbf{54.4} & \textbf{52.0} & \textbf{74.1} & \textbf{54.2} & 18.0 & \textbf{40.8} & \textbf{46.8} & \textbf{42.5} \\
\hline
Mantis-8B & 81.5 & \textbf{46.4} & 32.0 & 41.6 & 57.2 & 51.7 & 28.6 & 26.2 & 15.2 & 28.9 & 24.7 & 53.2 \\
Mantis-8B$^{\dag}$ & \textbf{81.6} & \textbf{46.4} & \textbf{32.3} & \textbf{42.2} & \textbf{57.5} & \textbf{52.0} & \textbf{37.4} & \textbf{26.9} & \textbf{15.9} & \textbf{29.7} & \textbf{27.5} & \textbf{54.8} \\
\hline
InternVL2-8B & 40.5 & 63.8 & 30.8 & 47.4 & 66.2 & 49.7 & 51.9 & 83.6 & \textbf{22.3} & 51.0 & 52.2 & 55.4 \\
InternVL2-8B$^{\dag}$ & \textbf{58.3} & \textbf{64.2} & \textbf{35.5} & \textbf{49.7} & \textbf{66.9} & \textbf{54.9} & \textbf{82.4} & \textbf{84.5} & 21.6 & \textbf{52.0} & \textbf{60.1} & \textbf{59.7} \\
\hline
Qwen-VL-7B & \textbf{39.9} & \textbf{3.2} & 4.8 & 13.3 & 24.0 & 17.0 & 32.2 & 22.0 & 19.1 & 15.7 & 22.3 & 28.5 \\
Qwen-VL-7B$^{\dag}$ & \textbf{39.9} & \textbf{3.2} & \textbf{5.0} & \textbf{14.2} & \textbf{26.5} & \textbf{17.8} & \textbf{38.1} & \textbf{22.6} & \textbf{20.8} & \textbf{16.5} & \textbf{24.5} & \textbf{29.6} \\
\hline
Idefics2-8B & \textbf{82.9} & 49.0 & \textbf{24.0} & 1.7 & 4.6 & 32.4 & \textbf{61.9} & 42.7 & 5.9 & 33.3 & 36.0 & 48.9 \\
Idefics2-8B$^{\dag}$ & 81.7 & \textbf{49.1} & 23.2 & \textbf{28.1} & \textbf{44.7} & \textbf{45.4} & 60.2 & \textbf{46.1} & \textbf{9.6} & \textbf{33.4} & \textbf{37.3} & \textbf{50.5} \\
\hline
LLaVA-OV-Qwen2-7B & 87.0 & 79.9 & 46.9 & \textbf{55.3} & 67.0 & 67.2 & 94.2 & \textbf{65.0} & \textbf{19.5} & 61.7 & 60.1 & 59.7 \\
LLaVA-OV-Qwen2-7B$^{\dag}$ & \textbf{88.1} & \textbf{83.6} & \textbf{47.0} & 55.0 & \textbf{67.3} & \textbf{68.2} & \textbf{96.3} & 62.9 & 19.4 & \textbf{62.5} & \textbf{60.3} & \textbf{62.4} \\
% Qwen2-VL-7B & 80.6 & 78.5 & 31.7 & 58.5 & 70.3 & 63.9 & 95.4 & 78.4 & 39.6 & 59.5 & 68.2 \\

% Qwen2-VL-7B\dag &  &  &  &  &  &  &  &  &  &  &  \\ 

\bottomrule
\end{tabular}}
\caption{Performance comparison by integrating our compression method (noted as ``\dag'') into different MLLMs tested on the MIBench and Mantis Eval. \textbf{Avg} indicates the average accuracy.}
\label{tab:res_on_mibench}
% \vspace{-2mm}
\end{table*}

\section{Experiment}
\subsection{Implementation details}
\subsubsection{Benchmark}
Our experiments are conducted on MuirBench \cite{wang2024muirbench}, MIBench \cite{liu2024mibench}, and Mantis-Eval \cite{jiang2024mantis}, all of which focus on MVQA scenarios. 
Mantis-Eval \cite{jiang2024mantis} is a carefully curated benchmark comprising $217$ questions, covering themes such as size perception and weight comparison. 
MuirBench \cite{wang2024muirbench}, specifically designed for MLLMs, includes $12$ tasks spanning $11,264$ images and $2,600$ multiple-choice questions. It features $10$ categories of image relationships, simulating human-like multi-source information integration capabilities. 
MIBench \cite{liu2024mibench} serves as a comprehensive, large-scale benchmark for evaluating the multi-image understanding capabilities of MLLMs. It contains $13$ sub-tasks and a total of $13,000$ image samples sourced from datasets such as NLVR2 \cite{suhr2019corpus}, MagicBrush \cite{zhang2024magicbrush}, and VrR-VG \cite{liang2019vrr}. MIBench also introduces Multi-Image Instruction (MII) and Multimodal Knowledge-Seeking (MKS) scenarios, designed to assess both the comparative reasoning abilities of MLLMs and their capacity to incorporate external knowledge when selecting from multiple answer choices. More details can be found in supplementary materials~\ref{sup_benchmark}.

\begin{table*}[!t]
\centering
\small
\resizebox{0.83\textwidth}{!}{
\begin{tabular}{l|ccccccccccccc}
\toprule
\multirow{2}{*}{\textbf{Model}} & \multicolumn{13}{c}{\textbf{MuirBech}} \\
 & \textbf{GU} &
\textbf{C} &
\textbf{AU} &
\textbf{VG} &
\textbf{ITM} &
\textbf{O} &
\textbf{SU} &
\textbf{DS} &
\textbf{CU} &
\textbf{DU} &
\textbf{AS} &
\textbf{VR} &
\textbf{Avg} \\
\midrule

LLaVA & \textbf{38.0} & 21.8 & 1.2 & 21.4 & 28.7 & 9.4 & 44.1 & 17.9 & 24.4 & \textbf{34.7} & 24.0 & \textbf{24.3} & 24.2 \\
\midrule
LLaVA + IR$_{R=0.1}$ & 24.0 & \textbf{26.1} & 27.4 & 21.4 & 29.3 & \textbf{20.3} & \textbf{50.0} & 20.0 & 19.2 & 27.6 & 25.0 & 20.6 & 26.0 \\
LLaVA + IR$_{R=0.3}$ & 26.0 & \textbf{26.1} & 29.3 & \textbf{22.6} & 31.0 & 15.6 & 48.9 & \textbf{20.3} & 21.8 & 27.4 & 26.5 & 19.9 & 26.3 \\
LLaVA + IR$_{R=0.5}$ & 32.0 & 23.1 & 28.1 & \textbf{22.6} & 30.8 & 14.1 & \textbf{50.0} & 18.8 & 24.4 & 30.7 & 27.6 & 21.6 & 27.0 \\
\midrule
LLaVA + IR$_{R=0.5}$ + CD$_{\lambda=7}$ & 30.0 & 22.7 & \textbf{31.1} & \textbf{22.6} & 31.0 & 12.5 & \textbf{50.0} & 18.2 & \textbf{25.6} & 30.2 & \textbf{29.1} & 21.6 & 27.1 \\
LLaVA + IR$_{R=0.5}$ + CD$_{\lambda=5}$ & 34.0 & 23.1 & \textbf{31.1} & \textbf{22.6} & 31.0 & 14.1 & \textbf{50.0} & 17.9 & \textbf{25.6} & 30.7 & \textbf{29.1} & 21.6 & \textbf{27.6} \\
LLaVA + IR$_{R=0.5}$ + CD$_{\lambda=3}$ & 33.0 & 21.8 & \textbf{31.1} & \textbf{22.6} & \textbf{31.5} & 14.1 & 50.0 & 18.2 & \textbf{25.6} & 29.9 & 27.0 & 21.6 & 27.2 \\
LLaVA + IR$_{R=0.5}$ + CD$_{\lambda=1}$ & 35.0 & 20.9 & 10.4 & \textbf{22.6} & 31.0 & 14.1 & 46.2 & 18.2 & 24.4 & 32.4 & 26.0 & 21.2 & 25.2 \\

\bottomrule
\end{tabular}}
\caption{Ablation study of method components and hyperparameters with LLaVA-v1.5 on MuirBench. \textbf{Avg} indicates the average accuracy across the benchmark.}
\label{tab:ablation_on_muirbench}
% \vspace{-2mm}
\end{table*}
\definecolor{mygray}{gray}{.92}

\begin{table*}[!t]
\centering
\small
\resizebox{0.83\textwidth}{!}{
\begin{tabular}{l|ccccccccccccc}
\toprule
\multirow{2}{*}{\textbf{Method}} & \multicolumn{13}{c}{\textbf{MuirBech}} \\
 & \textbf{GU} &
\textbf{C} &
\textbf{AU} &
\textbf{VG} &
\textbf{ITM} &
\textbf{O} &
\textbf{SU} &
\textbf{DS} &
\textbf{CU} &
\textbf{DU} &
\textbf{AS} &
\textbf{VR} &
\textbf{Avg} \\

\midrule
\rowcolor{gray!20}\multicolumn{14}{c}{\textbf{Upper Bound, 576 Tokens} (100\%)} \\
Vanilla \cite{liu2024visual} & 38.0 & 21.8 & 2.4 & 21.4 & 29.1 & 9.4 & 44.1 & 17.9 & 24.4 & 34.7 & 24.0 & \textbf{24.3} & 24.3 \\

\rowcolor{gray!20}\multicolumn{14}{c}{\textbf{Compressed to 192 tokens} (33.3\%)} \\
FastV \cite{chen2025fastv} & 34.0 & 15.8 & 1.8 & 21.4 & 28.7 & 12.5 & 36.0 & 17.1 & 21.8 & 30.9 & 23.0 & 19.9 & 21.9 \\
SparseVLM \cite{zhang2024sparsevlm} & 44.0 & 21.8 & 1.2 & 21.4 & 31.7 & 10.9 & 46.8 & 18.2 & 25.6 & 36.2 & 26.0 & 22.3 & 25.5 \\
FasterVLM \cite{zhang2024fastvlm} & 20.0 & 23.5 & 29.9 & \textbf{26.2} & \textbf{32.3} & 6.3 & \textbf{53.2} & 18.8 & 28.2 & 26.4 & 23.5 & 19.9 & 25.6 \\
VisionZip \cite{yang2024visionzip} & 20.0 & 25.2 & 28.7 & 22.6 & 31.3 & 6.3 & 51.6 & 19.7 & 26.9 & 25.4 & 22.5 & 19.9 & 25.0 \\

\rowcolor{gray!20}\multicolumn{14}{c}{\textbf{Compressed to 128 tokens} (22.2\%)} \\
FastV \cite{chen2025fastv} & 34.0 & 15.4 & 0.6 & 21.4 & 28.0 & 12.5 & 36.6 & 16.5 & 23.1 & 31.7 & 23.0 & 19.9 & 21.9 \\
SparseVLM \cite{zhang2024sparsevlm} & \textbf{46.0} & 22.7 & 4.3 & 21.4 & 32.0 & 9.4 & 46.8 & 18.2 & 25.6 & 36.2 & 27.0 & 23.6 & 26.1 \\
FasterVLM \cite{zhang2024fastvlm} & 20.0 & 23.1 & 29.9 & 23.8 & 29.7 & 7.8 & 52.7 & 19.1 & \textbf{32.1} & 25.9 & 23.0 & 19.9 & 25.6 \\
VisionZip \cite{yang2024visionzip} & 20.0 & 27.8 & 29.3 & 23.8 & 29.5 & 7.8 & \textbf{53.2} & 19.1 & 29.5 & 26.6 & 24.5 & 19.5 & 25.9 \\

\rowcolor{gray!20}\multicolumn{14}{c}{\textbf{Compressed to 64 tokens} (11.1\%)} \\
FastV \cite{chen2025fastv} & 32.0 & 15.4 & 4.9 & 23.8 & 27.8 & 12.5 & 36.0 & 16.5 & 24.4 & 31.7 & 23.0 & 19.9 & 22.3 \\
SparseVLM \cite{zhang2024sparsevlm} & 43.0 & 22.7 & 2.4 & 21.4 & 31.5 & 9.4 & 46.2 & 18.8 & 25.6 & \textbf{36.4} & 26.5 & 20.9 & 25.4 \\
FasterVLM \cite{zhang2024fastvlm} & 22.0 & 25.2 & 28.7 & 21.4 & 29.3 & 12.5 & 52.2 & \textbf{20.0} & 29.5 & 26.6 & 23.0 & 19.5 & 25.8 \\
VisionZip \cite{yang2024visionzip} & 20.0 & \textbf{28.6} & 28.7 & 21.4 & 28.9 & 10.9 & 52.7 & 18.5 & 29.5 & 25.9 & 23.5 & 19.9 & 25.7 \\

\rowcolor{gray!20}\multicolumn{14}{c}{\textbf{Adaptive Compression} (58.4\%)} \\
Ours & 32.0 & 23.1 & 28.1 & 22.6 & 30.8 & \textbf{14.1} & 50.0 & 18.8 & 24.4 & 30.7 & 27.6 & 21.6 & 27.0 \\

\rowcolor{gray!20}\multicolumn{14}{c}{\textbf{Adaptive Compression} (158.4\%)} \\
Ours$^{\ddagger}$ & 34.0 & 23.1 & \textbf{31.1} & 22.6 & 31.0 & \textbf{14.1} & 50.0 & 17.9 & 25.6 & 30.7 & \textbf{29.1} & 21.6 & \textbf{27.6} \\

% Qwen2-VL-7B & 80.6 & 78.5 & 31.7 & 58.5 & 70.3 & 63.9 & 95.4 & 78.4 & 39.6 & 59.5 & 68.2 \\

\bottomrule
\end{tabular}}
\caption{Performance change by integrating several compression methods into LLaVA-v1.5 with different compression sizes tested on the MuirBench. \textbf{Avg} indicates the average accuracy across the benchmark.}
\label{tab:compare_on_muirbench_diff_ratio}
% \vspace{-2mm}
\end{table*}

\subsubsection{Baselines}
The proposed method is theoretically compatible with most MLLMs, enabling performance improvements in multi-image scenarios. To validate its effectiveness, we evaluated eight mainstream MLLMs, including insertion-based models such as LLaVA-v1.5-7B \cite{liu2024improved}, DeepSeek-vl-7B~\cite{lu2024deepseek}, Mantis-8B \cite{jiang2024mantis}, InternVL2-8B \cite{chen2024internvl2} and LLaVA-OV-Qwen2-7B \cite{li2024llava_OV}, as well as query-learning-based models such as Qwen-VL-9.6B \cite{bai2023qwen} and Idefics2-8B \cite{laurenccon2024matters}. Our experiments followed the official settings recommended by the respective MLLMs, integrating our method to verify its adaptability and effectiveness. Supplementary materials~\ref{sup_model_details} provide the specific architectures of the MLLMs used.

\subsubsection{Evaluation}
Following the protocols of these benchmarks \cite{liu2024mibench, wang2024muirbench, jiang2024mantis}, we use the accuracy of MLLMs' predictions as the evaluation metric. For responses that do not directly provide a clear option, we leverage Llama3.2-11B \cite{dubey2024llama} to identify the option most semantically aligned with the model’s answer — a process shown to achieve strong performance across multiple reasoning benchmarks. For the evaluation on MIBench \cite{liu2024mibench}, we strictly follow its recommended testing procedure: the correct answer for each data instance is sequentially assigned to options ``A'', ``B'', ``C'', and ``D''. A sample is only considered correct if the model accurately answers all four permutations, effectively mitigating positional bias in option selection for MLLMs \cite{liu2024mibench, liu2025mmbench}.

\subsection{Compatibility with MLLMs}
We integrate the proposed method into several popular MLLMs and validate its effectiveness in preserving critical visual tokens through comprehensive evaluation on three multi-image benchmarks \cite{wang2024muirbench, liu2024mibench, jiang2024mantis}. As shown in Table \ref{tab:res_on_muirbench} and \ref{tab:res_on_mibench}, all evaluated multimodal large models achieve significant improvements in average accuracy across the benchmarks when equipped with our method.  By enhancing the models' focus on crucial visual tokens, our approach consistently facilitates more accurate reasoning and problem-solving, demonstrating both its effectiveness and strong generalizability. Notably, the extent of performance improvement varies across different model architectures.

\begin{figure*}
    \centering
    \includegraphics[width=0.912\linewidth]{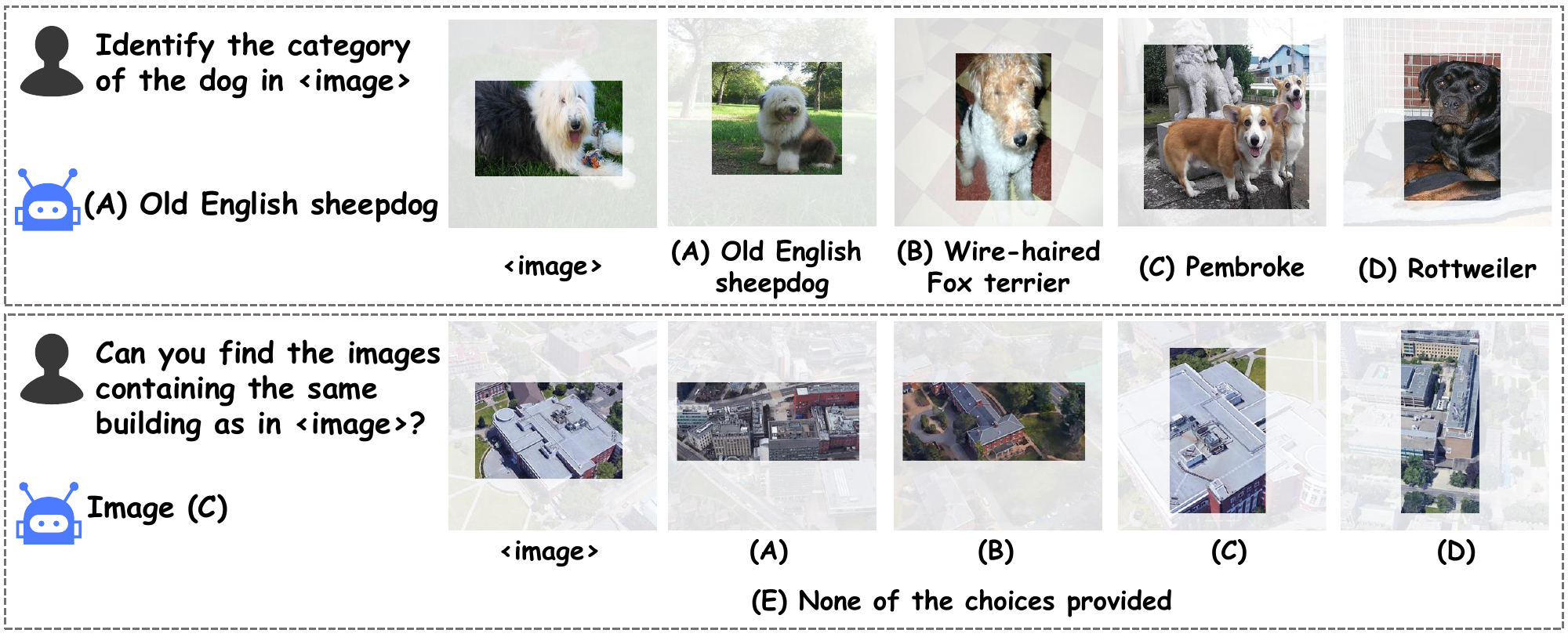}
    \caption{Two examples illustrating the QA results produced by our method, where critical visual regions are extracted using our adaptive visual anchoring strategy. These regions closely align with the given questions, facilitating accurate answer inference by the MLLM.}
    \label{fig:visualization}
\end{figure*}

For insertion-based methods (LLaVA \cite{liu2024visual}, DeepSeek-VL \cite{lu2024deepseek}, and Mantis \cite{jiang2024mantis}), we observe substantial accuracy improvements, particularly in MuirBench subtasks such as Image-Text Matching (ITM) and Difference Spotting (DS), as well as in MIBench tasks like Fine-grained Visual Recognition (FVR) and Text-linked Visual Knowledge (TVK). These tasks typically rely on text descriptions that are highly correlated with visual content (e.g., the precise textual descriptions provided for each image in FVR, as illustrated in Figure \ref{fig:architecture}). In particular, the original LLaVA-v1.5 \cite{liu2024visual} and DeepSeek-VL \cite{lu2024deepseek} fail to follow the instructions to generate valid results in Temporal Reasoning (TR) and Logical Reasoning (LR) tasks, primarily because the sequences of input tokens exceed their optimal context length. Our method effectively alleviates this issue by significantly reducing the number of visual tokens.

For query-based MLLMs such as Qwen-VL \cite{bai2023qwen} and Idefics2~\cite{laurenccon2024matters}, notable performance gains are observed in tasks such as Counting (C), Diagram Understanding (DU), Temporal Reasoning (TR), and Logical Reasoning (LR). However, limited improvements are observed in tasks requiring fine-grained visual discrimination (e.g., Geometric Comparison (GC), Shape Discrimination (SD), and Visual Recognition (VR)), indicating that the learnable queries may struggle to capture subtle inter-image differences when critical regions are not properly selected.

\subsection{Ablation study}
As shown in Table \ref{tab:ablation_on_muirbench}, we conduct an ablation study on the components and key parameters of our method using MuirBench \cite{wang2024muirbench}, with LLaVA \cite{liu2024visual} as the base model. After integrating Implicit Retrieval (IR), we observe significant performance improvements in most tasks, such as Action Understanding (AU) and Ordering (O), indicating that these tasks are particularly affected by the obstruction of critical visual tokens. However, for tasks that rely on global visual perception, such as Geographic Understanding (GU), answering questions typically requires interpreting the entire map’s semantic content. In these cases, aggressive token pruning may lead to semantic loss. This issue is partially alleviated by proposed Collaborative Decoding (CD), which integrates the original unpruned features, enabling further gains in tasks such as Action Understanding (AU) and Cartoon Understanding (CU). We further explore the impact of the collaboration coefficient $\lambda$, finding that when $\lambda = 5$, the balance between global decoding and decoding focused on critical visual regions achieves optimal alignment, resulting in the highest average accuracy across tasks. This highlights that excessive collaborative strength can introduce noise from redundant visual content in the original feature maps, ultimately degrading performance.

\subsection{Comparison with advanced methods}
We applied several mainstream visual token compression algorithms \cite{chen2025fastv, zhang2024sparsevlm, yang2024visionzip, zhang2024fastvlm} to multi-image scenarios and conducted experiments on MuirBench \cite{wang2024muirbench}, following the experimental settings reported in their respective papers. Each method was tested with varying numbers of retained visual tokens. As shown in Table \ref{tab:compare_on_muirbench_diff_ratio}, applying compression at different levels consistently improves average accuracy compared to the vanilla model, further confirming that redundant visual tokens negatively impact the ability of MLLMs to understand tasks effectively. 
Compared to methods such as FastV~\cite{chen2025fastv}, VisionZip~\cite{yang2024visionzip} and FasterVLM~\cite{zhang2024fastvlm}, which compress visual tokens based solely on intra-visual attention scores, SparseVLM \cite{zhang2024sparsevlm} leverages aligned cross-modal textual information to prune redundant tokens, achieving superior performance gains on most tasks. However, these methods excessively select sparse visual tokens, disrupting semantic continuity and substantially degrading performance in spatial reasoning tasks, such as Ordering (O) and Attribute Similarity (AS), compared to the vanilla model. In contrast, our proposed method dynamically performs collaborative decoding based on redundancy rates after implicit retrieval, achieving a superior overall accuracy. 
Although our method does not show advantages in terms of compression ratio, it achieves significant performance improvements compared to methods without compression or with fixed compression sizes. This finding further demonstrates that our strategy can effectively filter out irrelevant information while preserving key visual details, thereby preventing the loss of crucial content.

\subsection{Qualitative visualization}
Fig.~\ref{fig:visualization} presents two examples illustrating the visualization results of our approach. Given a textual question, our method accurately captures visual regions in each image that are highly semantically relevant to the question while filtering out redundant visual tokens. Consequently, the compressed visual tokens are fed into LLM, enabling a precise alignment between visual representation and the question, thereby facilitating the generation of correct answers.

\section{Conclusion}
This paper introduces a novel strategy, ``Adaptive Visual Anchoring'', marking the first exploration of visual token compression for MVQA. We conduct an in-depth analysis of the limitations of existing visual token compression methods in handling MVQA, particularly issues related to visual fragmentation and fixed compression rates across multiple images. The proposed method adaptively captures critical visual regions relevant to the given question, providing MLLMs with more precise visual representations to generate accurate answers. Furthermore, our approach can be seamlessly integrated into existing MLLMs, enhancing both efficiency and accuracy in MVQA. Extensive experiments validate the effectiveness of our method.

\clearpage
\section*{Acknowledgements}
\vspace{-5pt}
This work was partially supported by National Natural Science Foundation of China (No.U21A20518, No.U23A20341, No.62272157).

{
    \small
    \bibliographystyle{ieeenat_fullname}
    % \bibliography{main}
    \bibliography{tex_content/reference}
}

\clearpage
\maketitlesupplementary
\appendix
\section{Benchmark details}
\label{sup_benchmark}
\subsection{MuirBench}
\label{sup_muirbench}
MuirBench is designed for multi-image understanding, comprising $11,264$ images and $2,600$ multiple-choice questions, with an average of $4.3$ images per instance.
MuirBench evaluates models across $12$ key tasks, each representing $2.5\%$ to $17.8\%$ of the dataset:
\begin{itemize}
    \item Geographic Understanding (GU): Reasoning over maps and geographic features (e.g., \textit{``Among these map images, which one depicts overlapping geographic regions like $<img1>$?''}).
    \item Counting (C): Quantifying specific objects across multiple images (e.g., \textit{``How many vases have a painted design all over in the images?''}).
    \item Action Understanding (AU): Matching sequential images to actions (e.g., \textit{``What is the action displayed in the video?''}). 
    \item Visual Grounding (VG): Locating specific objects and extracting relevant information (e.g., \textit{``This is the McDonald's my sister bought $<img1>$. This is the McDonald's \$1 \$2 \$3 Dollar Menu $<img2>$. Could you please tell me how much my sister spent on this McDonald's?''}).
    \item Image-Text Matching (ITM): Associating text snippets with corresponding images (e.g., \textit{``Which images has $1$ apple and $5$ bananas?''}).
    \item Ordering (O): Arranging images based on textual descriptions (e.g., \textit{``The baby attempts to take off the clothes. What is the correct order of images according to the given context?''}).
    \item Scene Understanding (SU): Analyzing multi-view scenes from surveillance images (e.g., \textit{``What's the color of the car parked behind the black van in the given images?''}).
    \item Difference Spotting (DS): Identifying differences between images (e.g., \textit{``Can you determine which slide serves a different function compared to the others?''}).
    \item Cartoon Understanding (CU): Interpreting stories conveyed in cartoon images (e.g., \textit{``What is the main content of this comic strip?''}).
    \item Diagram Understanding (DU): Extracting information from diagram images (e.g., \textit{``Which object is below the bed?''}).
    \item Attribute Similarity (AS): Identifying a specific attribute across multiple images (e.g., \textit{``Which of the following images shares the same scene with $<img1>$ but contains the object potted plant?''}).
    \item Visual Retrieval (VR): Identifying images containing the same building (e.g., \textit{``Can you find the images containing the same building as in $<img1>$?''}).
\end{itemize}

\subsection{MIBench}
\label{sup_mibench}
In MIBench, multi-image inputs are categorized into three scenarios: Multi-Image Instruction (MII), Multimodal Knowledge-Seeking (MKS), and Multimodal In-Context Learning (MIC).
\begin{itemize}
\item MII involves perception, comparison, and reasoning across multiple images (e.g., \textit{“Do the two images show the same number of cats?”}).
\item MKS assesses an MLLM’s ability to retrieve relevant information from external knowledge provided in an interleaved image-text format. Unlike MII, MKS questions may focus on a single image or be independent of visual content.
\item MIC evaluates MLLMs’ ability to answer visual questions with the aid of multimodal demonstrations (i.e., examples).
\end{itemize}

\subsubsection{Multi-Image Instruction (MII)}
Based on the semantic types of instructions, MII is further divided into $5$ tasks:
\begin{itemize}
    \item General Comparison (GC): Evaluates the model’s understanding of individual images, including aspects like scene, attributes, and location, and its ability to compare these images (e.g., \textit{``Can the given sentence accurately illustrate what's in these two images? Two dogs are lying in the grass in each of the images.''}).
    \item Subtle Difference (SD): Assesses fine-grained perception to detect minor differences between similar images (e.g., \textit{``What are the differences between image 1 and image 2?''}).
    \item Visual Referring (VR): Tests the model’s ability to understand object relationships based on referring expressions (e.g., \textit{``Based on image 1, what is the relationship between image 2 and image 3?''}).
    \item Temporal Reasoning (TR): Measures comprehension of temporal relationships in consecutive images (e.g., \textit{``What action do these images show?''}).
    \item Logical Reasoning (LR): Requires causal reasoning about objects or events depicted in images (e.g., \textit{``Why did the boy in black extended his hands after the boy in white extended his hands?''}).
\end{itemize}

\subsubsection{Multimodal Knowledge-Seeking (MKS)}
Based on the form of external knowledge, MKS is divided into $4$ tasks:
\begin{itemize}
    \item Fine-grained Visual Recognition (FVR): Evaluates the model’s ability to recognize objects in a query image using multiple reference images, requiring an understanding of image-label correspondence and similarity linking (e.g., \textit{``Look at the dog pictures presented above and tell me which type of dog is represented in this image.''}).
    \item Text-Rich Images (TRI) VQA: Assesses the model’s ability to extract relevant information from text - heavy images, which represent a common real-world scenario involving tasks like reading slides and documents (e.g., \textit{`` What is the population of the country where the cabinet is named `Kabinet Kerja'?''}).
    \item Vision-linked Textual Knowledge (VTK): Tests the model’s ability to link query images with relevant external knowledge (e.g., Wikipedia) and extract useful information from corresponding text (e.g., \textit{``Which city or region does this building locate in?''}).
    \item Text-linked Visual Knowledge (TVK): Evaluates the model’s capability to answer text-only questions about the visual attributes of specific objects when given interleaved image-text knowledge (e.g., \textit{``At the victory ceremony for Boxing at the 2018 Summer Youth Olympics how many medalists were holding their hand over their heart?''}).
\end{itemize}

\subsubsection{Multimodal In-Context Learning (MIC)}
In-context learning allows LLMs to improve performance when provided with a series of demonstrations. Recent studies introduce a more fine-grained assessment by dividing MIC into $4$ distinct tasks:
\begin{itemize}
    \item Close-ended VQA: Requires the model to select answers from a predefined set provided via multimodal demos, assessing its ability to learn image-label mappings.
    \item Open-ended VQA: Evaluates the model’s ability to infer task patterns from demos when answers fall outside the predefined set.
    \item Hallucination Mitigation: Investigates the impact of MIC on hallucination.
    \item Demo-based Task Learning: Tests the model’s ability to rapidly adapt to new tasks with few-shot demonstrations by removing explicit task instructions and presenting demos in a structured format (e.g., \textit{“rabbit: $3$”}).
\end{itemize}

\subsection{Mantis-Instruct}
\label{sup_mantis}
Mantis-Instruct, the first multi-image instruction-tuning dataset, comprising $721$K instances across $14$ subsets, designed to cover all essential multi-image skills.

\begin{figure*}
    \centering
    \includegraphics[width=\linewidth]{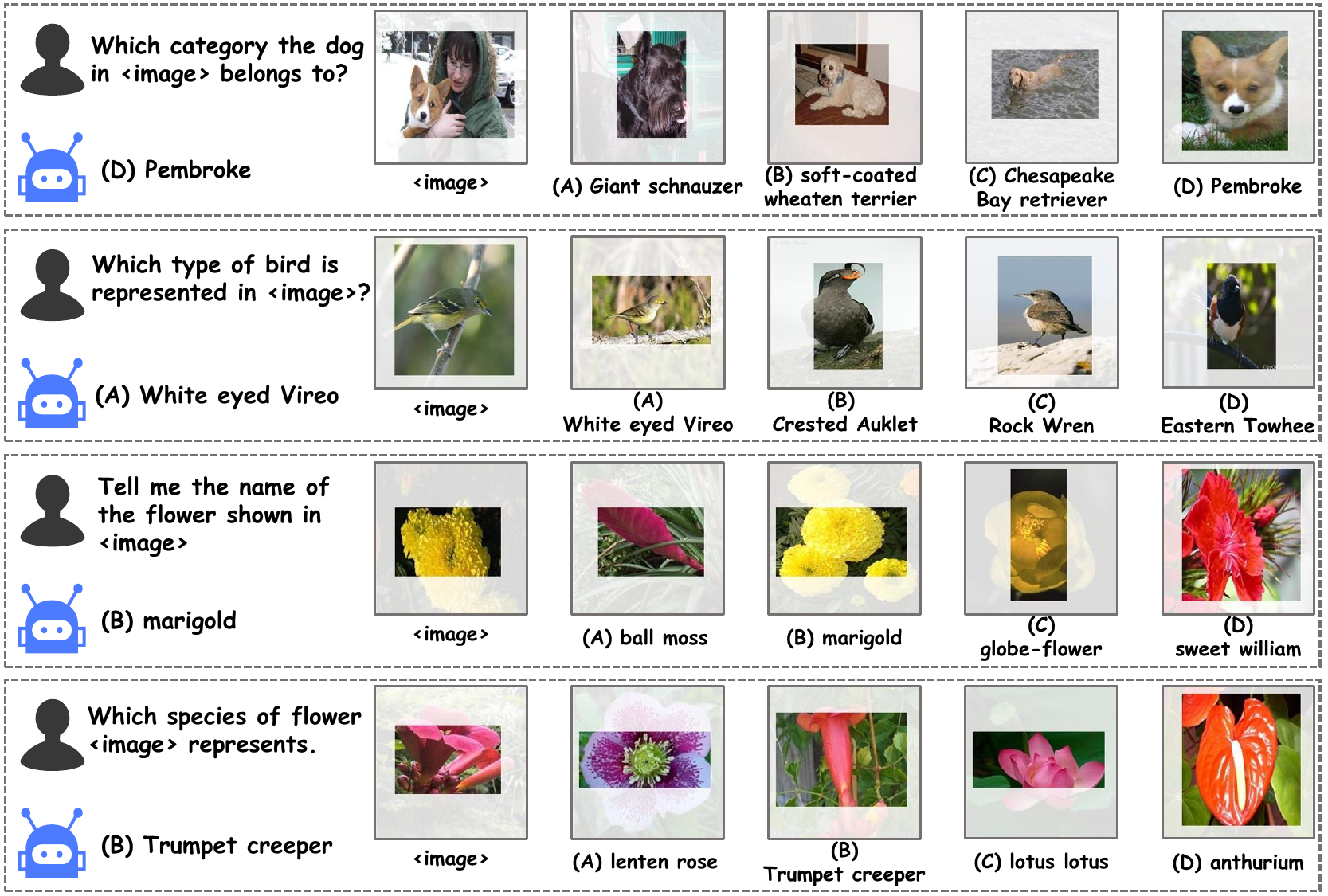}
    \caption{More Qualitative visualization results.}
    \label{fig:sup_visualization}
\end{figure*}

$10$ subsets are sourced from existing datasets:
\begin{itemize}
\item Reasoning: NLVR2, IconQA.
\item Comparison: DreamSim, Birds-to-Words.
\item Temporal Understanding: NExT-QA, STAR.
\end{itemize}
4 newly curated subsets:
\begin{itemize}
\item Coreference Resolution: LLaVA-665k-multi, LRV-multi.
\item Expanded Reasoning: Contrast-Caption, Multi-VQA.
\end{itemize}
To enhance instruction formatting, interleaving image placeholders are inserted into text based on various heuristics.

\section{The impact of black image injection on distribution shift}
\label{sup_the_impact_of_black_images}
We added redundant black blank images to the visual input to explore the impact of visual redundancy on MLLMs in the MVQA task. However, this impact might also stem from the distribution shift of visual input, i.e., the pure black images themselves could affect the output responses of MLLMs. Thus, we designed an exploratory experiment to demonstrate the true cause of the performance drop in Fig. \ref{fig:analysis}(a).
\begin{table}[!t]
\centering
\setlength{\tabcolsep}{8pt}
\resizebox{0.47\textwidth}{!}{
    \begin{tabular}{l|c|ccc}
    \toprule 
    \multirow{3}{*}{\textbf{MLLM}} & \multicolumn{4}{c}{\textbf{Accuracy}} \\
    \cline{2-5}
       & \multirow{2}{*}{\textbf{Vanilla}} & \multicolumn{3}{c}{\textbf{Replaced visual inputs}}  \\
    \cline{3-5}
       &    &   \textbf{1}   &   \textbf{2}   &   \textbf{3}   \\
    \cline{1-5}
    Mantis-8B & 28.6 & 62.8 & 74.2 & 98.9 \\
    Qwen-VL-9.6B & 32.2 & 37.8 & 48.1 & 79.2 \\
    \bottomrule 
    \end{tabular}}
\caption{Accuracy of MLLMs on the FVR task under different numbers of original visual input replacements, where ``Vanilla'' denotes $0$ replaced visual inputs, and columns $1$–$3$ represent $1$, $2$, and $3$ replaced visual inputs, respectively.}
\label{tab:tab6_replaceExp}
\end{table}
As shown in Tab.~\ref{tab:tab6_replaceExp}, we replaced the original visual inputs (except the correct answer) in the FVR task with $1$-$3$ pure black images, respectively. Notably, the accuracy of MLLMs in question answering gradually increased as more visual inputs were substituted. This improvement occurs because the reduction of potentially confusing images, coupled with the MLLMs' ability to recognize these blank black images, facilitates the selection of the correct answers. This experiment serves to further validate the conclusion drawn in Sec \ref{submergence}.

\section{Model details}
\label{sup_model_details}
\begin{itemize}
    \item LLaVA-v1.5-7B \cite{liu2024visual}: CLIP ViT-L/14 \cite{radford2021clip} serves as the vision encoder and Vicuna-v1.5-7B \cite{zheng2023vicuna} as the LLM.
    \item DeepSeek-vl-7B \cite{lu2024deepseek}: SAM-B \cite{kirillov2023sam} \& SigLIP-L \cite{zhai2023siglip} serve as the vision encoder and DeepSeek-7B \cite{bi2024deepseek_llm} as the LLM.
    \item Mantis-8B \cite{jiang2024mantis}: SigLIP SoViT-400M/14~\footnote{https://huggingface.co/google/siglip-so400m-patch14-384\label{siglip_sovit}} serves as the vision encoder and Llama3-8B \cite{dubey2024llama} as the LLM.
    \item InternVL2-8B \cite{chen2024internvl2}: InternViT-300M-448px \cite{chen2024internvl} serves as the vision encoder and InterLM2.5-7B \cite{cai2024internlm2} as the LLM.
    \item Qwen-VL-9.6B \cite{bai2023qwen}: CLIP ViT-G/14 \cite{radford2021clip} serves as the vision encoder and Qwen-7B \cite{bai2023qwen_lm} as the LLM.
    \item Idefics2-8B \cite{laurenccon2024matters}: SigLIP-L \cite{zhai2023siglip} serves as the vision encoder and Mistral-7B \cite{Mistral7B} as the LLM.
    \item LLaVA-OV-Qwen2-7B \cite{li2024llava_OV}: SigLIP SoViT-400M/14~\textsuperscript{\ref{siglip_sovit}} serves as the vision encoder and Qwen2-7B\cite{team2024qwen2_lm} as the LLM.
\end{itemize}

\section{Ablation of text selection strategy}
\label{sup_text_selection}
\begin{table}[h!]
\centering
\vskip-2.2ex
\setlength{\tabcolsep}{8pt}
\resizebox{0.47\textwidth}{!}{
    \begin{tabular}{l|c|c}
    \toprule
    \multirow{2}{*}{\textbf{Strategy}} & \multicolumn{2}{c}{\textbf{FVR task of MIBench}} \\
    \cline{2-3}
     & \textbf{Accuracy} & \textbf{Compression ratio} \\
     \cline{1-3}
     Vanilla & 29.2 & 0.0\% \\
     Question-based & 35.8 & 54.2\% \\
     Caption-based & \textbf{36.8} & \textbf{57.6\%} \\
    \bottomrule
    \end{tabular}}
\caption{Comparison of different text selection strategies. ``Vanilla'' denotes the original LLaVA-v1.5-7B, while ``Question-based'' and ``Caption-based'' represent the results of obtaining response maps using question texts and corresponding image captions, respectively.}
\label{tab:tab7_text_select}
\end{table}
Captions offer object-centric textual descriptions, which are typically more precise than question texts, and facilitate the accurate extraction of critical visual tokens. However, such captions are not always available. In the visual anchoring process, the response map is calculated exclusively from the image and question text when captions are absent. We compare caption-based and question-based strategies on the FVR task as shown in Tab.~\ref{tab:tab7_text_select}, where the latter replaces captions with questions to extract critical visual regions. Although question-based strategies use coarser text than captions, they effectively mitigate the submergence of critical visual tokens by visual redundancy. This approach does exhibit a slight performance decline, as question texts have less direct relevance to visual inputs than captions, leading to less accurate anchoring.

\end{document}